\documentclass{article}

\usepackage{fullpage}
\usepackage{subcaption}
\usepackage{hyperref}       
\usepackage{amsfonts}
\usepackage{amsmath}
\usepackage{graphicx}
\usepackage{times}
\usepackage[toc,page]{appendix}
\usepackage{natbib}
\usepackage{amsthm}
\usepackage[usenames, dvipsnames]{color}
\usepackage{microtype}

\usepackage{authblk}

\newcommand{\rulesep}{\unskip\ \vrule\ }

\newtheorem{theorem}{Theorem}[section]

\newtheorem{lemma}[theorem]{Lemma}

\begin{document}

\title{Weighted Tensor Decomposition for \\ Learning Latent Variables with Partial Data }
\author[1]{\textbf{Omer Gottesman}\thanks{\href{mailto:gottesman@fas.harvard.edu}{gottesman@fas.harvard.edu}}}
\author[1]{\textbf{Weiwei Pan}}
\author[1]{\textbf{Finale Doshi-Velez}\thanks{\href{mailto:finale@seas.harvard.edu}{finale@seas.harvard.edu}}}
\affil[1]{Paulson School of Engineering and Applied Sciences, Harvard University}
\date{}
\maketitle

\begin{abstract}
Tensor decomposition methods are popular tools for learning latent
variables given only lower-order moments of the data. However, the
standard assumption is that we have sufficient data to estimate these
moments to high accuracy. In this work, we consider the case in which
certain dimensions of the data are not always observed---common in
applied settings, where not all measurements may be taken for all
observations---resulting in moment estimates of varying quality.  We
derive a weighted tensor decomposition approach that is
computationally as efficient as the non-weighted approach, and
demonstrate that it outperforms methods that do not appropriately
leverage these less-observed dimensions.
\end{abstract}

\section{Introduction}
\label{sec:intro}
Tensor decomposition methods decompose low-order moments of observed
data to infer the latent variables from which the data is generated
\citep{anandkumar2014tensor,arora2012learning}.  They have recently
gained popularity because they require only moments of the data and,
in the absence of model-mismatch, come with global optimality
guarantees.  Of course, these guarantees require that the moments are
well-estimated---which corresponds to the amount of available
data. Previous works have investigated the dependence of the inference
quality on the amount of collected data (and the related uncertainty
in the moment estimates) \citep{anandkumar2014tensor, kolda2009tensor,
  podosinnikova2015rethinking}.  However, these works typically assume
that every dimension is observed in every datum.

In this work, we consider the scenario in which some dimensions are
always observed, while others may be available for only a few
observations.  Such situations naturally arise in real world
applications. In clinical settings, for example, easily obtained
measurements such as temperature and blood pressure may be available
for all patients, but the results of a blood test may be available for
only a small number of patients.  Similarly, billing codes may be
available for all patients, while information extracted from
unstructured notes only for a subset.  In biology applications, one
may want to combine counts from whole genome sequencing from a smaller
study with counts of the subset of SNPs from a larger population.

Let us suppose that we are only interested in latent variable models over
the dimensions that are completely observed. In the medical setting
example we presented earlier, that would correspond to learning a
model to explain only the patients' temperature and blood pressure
data. In such a case, would using the data on other vitals be helpful
or detrimental to learning?  We demonstrate that depending on the
frequency at which a dimension is missing in the data, including its
associated estimated moments may help or hinder the estimation of
latent variable model parameters for the rest of the model.

Of course, the more interesting question is whether we can do better
than sometimes ignoring dimensions with missingness---that is, can the
lower quality moment estimates from these incompletely-observed
dimensions still be used to improve our inference about the structure
of the complete dimensions?  We introduce a weighted tensor
decomposition method (WTPM) which weights elements in the moment
estimates proportionally to the frequency at which they are
observed. It has identical computation requirements to the unweighted
approach, and performs as well as or better than both unweighted
approaches, i.e. including or ignoring the dimensions which are
sometimes missing. We present experimental results for two commonly
used latent variable models: Gaussian mixtures and the Gamma-Poisson
model.

%

\section{Background and Setting}

\subsection{Tensor decomposition methods}

The principle at the heart of learning latent variable models via
tensor decomposition is simple: the moments of many such models can be
written as a tensor decomposition of the models parameters in the form
\begin{align}
\textbf{S} & = \sum_k s_k \textbf{a}_k \textbf{a}_k^T, \label{eq:S_decomposition}\\
\textbf{T} & = \sum_k t_k \textbf{a}_k \otimes \textbf{a}_k \otimes \textbf{a}_k. \label{eq:T_decomposition}
\end{align}
Here, $\textbf{a}_k$ denotes columns of the parameter matrix $\textbf{A}\in \mathbb{R}^{D\times K}$, representing the $k^{th}$ latent parameter in a $D$ dimensional space. $\textbf{A}$ can then be inferred from the data - denoted for a dataset of $N$ samples by the matrix $\textbf{X}\in\mathbb{R}^{D\times N}$ - by computing the empirical estimates
of the moments $\hat{\textbf{S}}(\textbf{X})$ and $\hat{\textbf{T}}(\textbf{X})$, and calculating their decomposition. The procedure is
quite general and can be applied whenever the expression for the
empirical estimates of the moments can be derived and computed from
the data. In this work we present results for mixtures of spherical Gaussians \citep{hsu2013learning} and the Gamma-Poisson (GP) model \citep{podosinnikova2015rethinking}. A description of the generative models and the form of the empirical estimates can be found in appendix \ref{appendix:empirical_moments}.

Several methods exist to estimate the decomposition of the tensor
estimates
\citep{hyv1999fast,cardoso1993blind,anandkumar2014tensor}. In this
work, we use the method described in \citet{anandkumar2014tensor},
which provides a polytime algorithm to learn the parameters from the
moment estimates. The method consists of two stages. In the first
stage the eigenstructure of $\hat{\textbf{S}}$ is used to compute a
whitening matrix, $\textbf{W}$, which is used to reduce
$\hat{\textbf{T}}$ to a $K\times K\times K$ tensor with an orthogonal
decomposition, $\hat{\textbf{T}}_c$. In the second stage, the tensor
power method (TPM) - the tensor analog of the matrix power method - is
used to calculate the eigenvectors of $\hat{\textbf{T}}_c$, which can
be transformed into the columns of the parameter matrix
$\hat{\textbf{A}}$ using $\textbf{W}$ once again. (For more details
see \citet{anandkumar2014tensor}). We emphasize that our results are
not specific to this choice of decomposition method.

\begin{figure}[t] 
\centering
\includegraphics[width=0.5\linewidth]{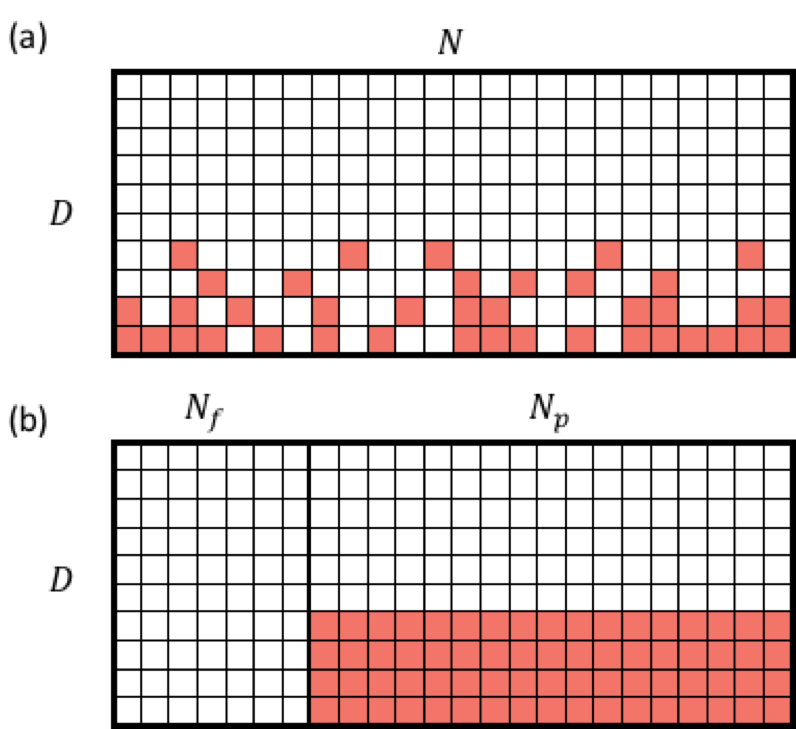}
\caption{(a) Basic
  uncorrelated structure of missing data. Colored elements represent missing data
  elements which are not observed.  Specifically, every dimension $d^{th}$ in a
  sampled data point $x_n$ (column $n$ in $\textbf{X}$) is present
  with some probability $p_d$, independently of the missingness of all
  other elements in the data. (b) Correlated missing data structure in the toy
  example.}
\label{data_structure}
\end{figure}

\subsection{Missingness}
In the following, we assume that there exist a certain set of
\emph{complete dimensions} that are always recorded in every
observation.  It is the relationships between these dimensions that we are most interested in recovering. The remaining
\emph{incomplete dimensions} are missing completely-at-random with
probability $(1-p_d)$. That is, the probability that one of these
dimensions is not recorded in observation $n$ is independent of all
other dimensions in $\textbf{x}_n$, the data set $\textbf{X}$, and the
specific value of $x_{nd}$. A schematic illustration of such a
dataset is presented in Figure~\ref{data_structure} (a). To calculate $p_d$ empirically we simply count the number of samples in which a given dimension is observed, and divide by the number of samples, $N$.

\section{The Value of Incomplete Data: Motivating Example}
\label{sec:motivation}
In Section~\ref{sec:intro}, we asked a simple question: if some of our
observations are incomplete---that is, we have observed only a few of
the possible dimensions---then is it better to just discard them?  Or
are there instances in which the additional poorly-estimated moments
are valuable for recovering parameters associated with the complete
dimensions? In this section, we demonstrate that the answer depends on \emph{how}
poorly estimated those additional moments are---there are
regimes in which the additional moments only add noise, and regimes in
which they assist with the overall parameter recovery.

We consider a very simple missingness pattern for the purpose of
illustration, shown schematically in Figure~\ref{data_structure} (b).
Specifically, we consider a data set that consists of $N_f$ full
observations, for which all dimensions are observed, and $N_p$ partial
observations, for which some dimensions are never observed. (Note that
this structure is not missing-at-random, which we will require for our
derivations of Lemma \ref{lem:optimal_weights}, but makes it simpler to illustrate the idea.)

Our metric for the reconstruction error of the parameter matrix $\textbf{A}$ incorporates
\emph{only} the complete dimensions:
\begin{equation}
\varepsilon_c \equiv \sum_k \frac{\hat{\textbf{a}}_{c,k}\cdot\textbf{a}_{c,k}}{||\hat{\textbf{a}}_{c,k}||\cdot||\textbf{a}_{c,k}||},
\label{eqn:error} 
\end{equation}
where $\textbf{a}_{c,k}$ is the vector comprised only of the
\emph{complete} elements in the $k^{th}$ latent parameter. We measure
the reconstruction error as the sum of angles in high dimensional
space between the true and inferred parameters in order to avoid
ambiguities due to scaling and normalization. (In particular, the
columns of $\textbf{A}$ in the GP model are normalized to sum to one, so it is
necessary to choose a measure of error which is independent of
normalization.)

In our illustration, we use the GP model and consider the recovery of
a single topics matrix $\hat{\textbf{A}}\in\mathbb{R}^{10\times4}$,
shown in Figure~\ref{features_leading_to_concept} (c) (In
Appendix~\ref{appendix:insensitivity_to_topics_structure}, we
demonstrate that our results also hold across a large number of
randomly generated topic matrices $\textbf{A}$).

\begin{figure}[t]
\centering
\begin{subfigure}{\linewidth}
\centering
  \includegraphics[width=\linewidth]{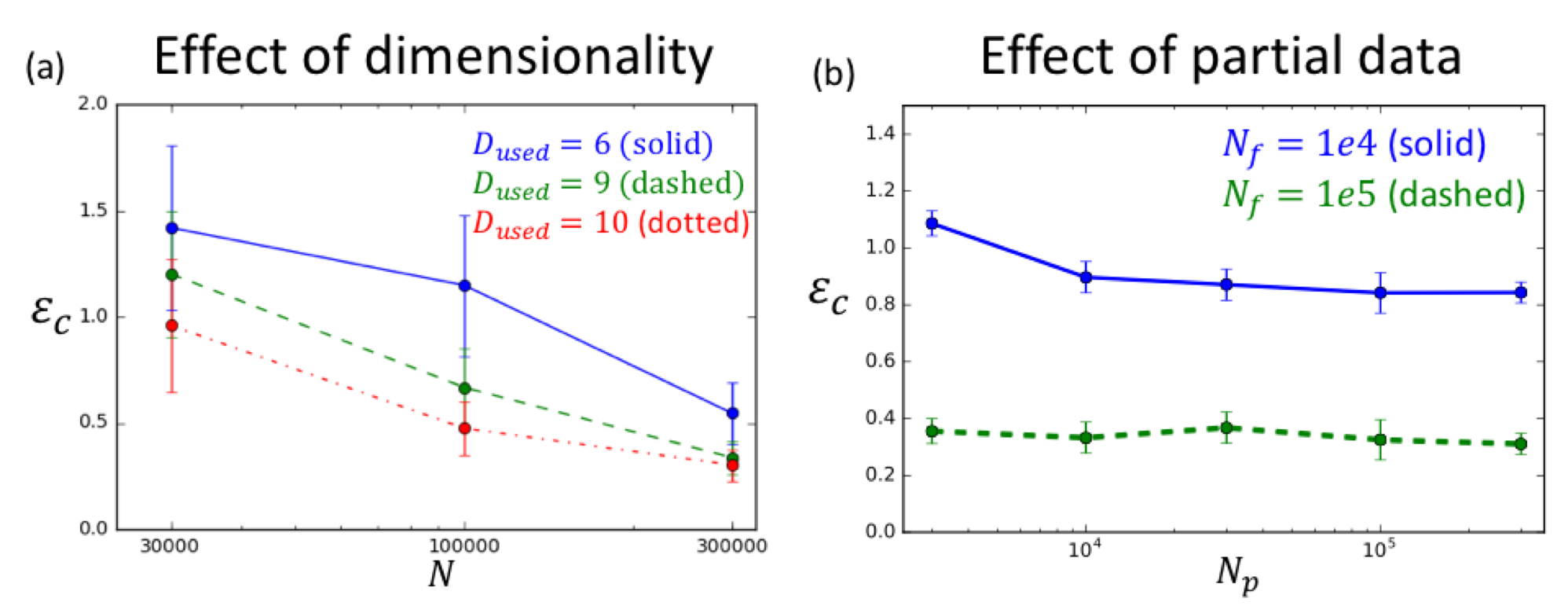}
\end{subfigure}

\begin{subfigure}{\linewidth}
\includegraphics[width=\linewidth]{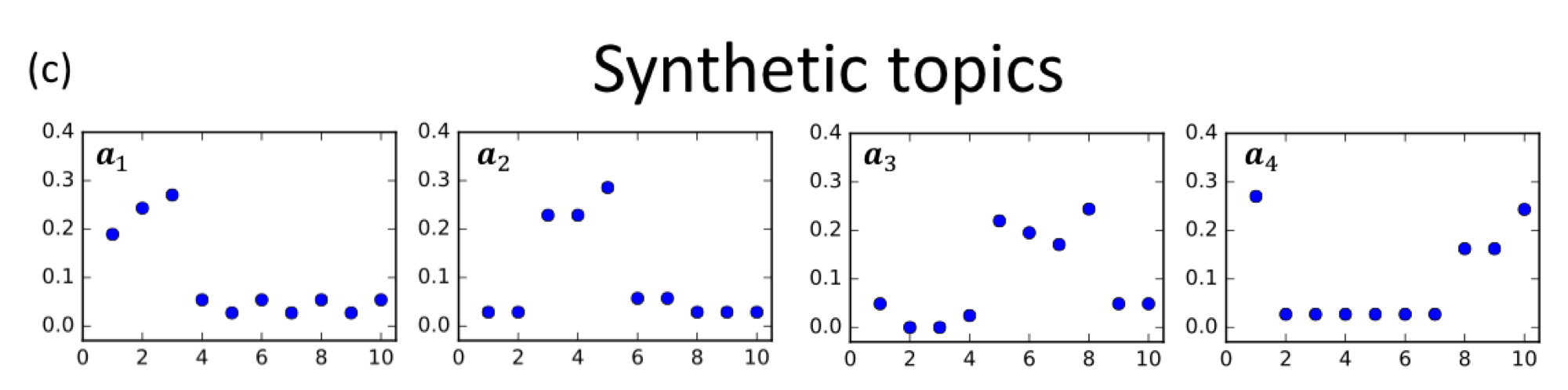}
\centering  
\end{subfigure}
\caption{Effect of dimensionality and additional partially-observed
  data. In (a), the additional correlations by observing additional
  dimensions reduces the reconstruction error on the six dimensions of
  interest, for any number of samples $N$.  In (b), increasing the
  number of partial samples $(N_p)$, for which 4 out of $D=10$
  elements are missing, has a limited effect on reconstruction error.
  Each data point is an average over 20 runs in which the data was
  sampled independently, and error-bars represent the standard
  deviation of the inference errors for each run. (c) Structure of $\textbf{A}$ in the synthetic GP example. The topics elements
  $a_{dk}$ are plotted as a function of $d$.}
\label{features_leading_to_concept}
\end{figure}

\begin{figure}[t]
\centering
\includegraphics[width=0.5\linewidth]{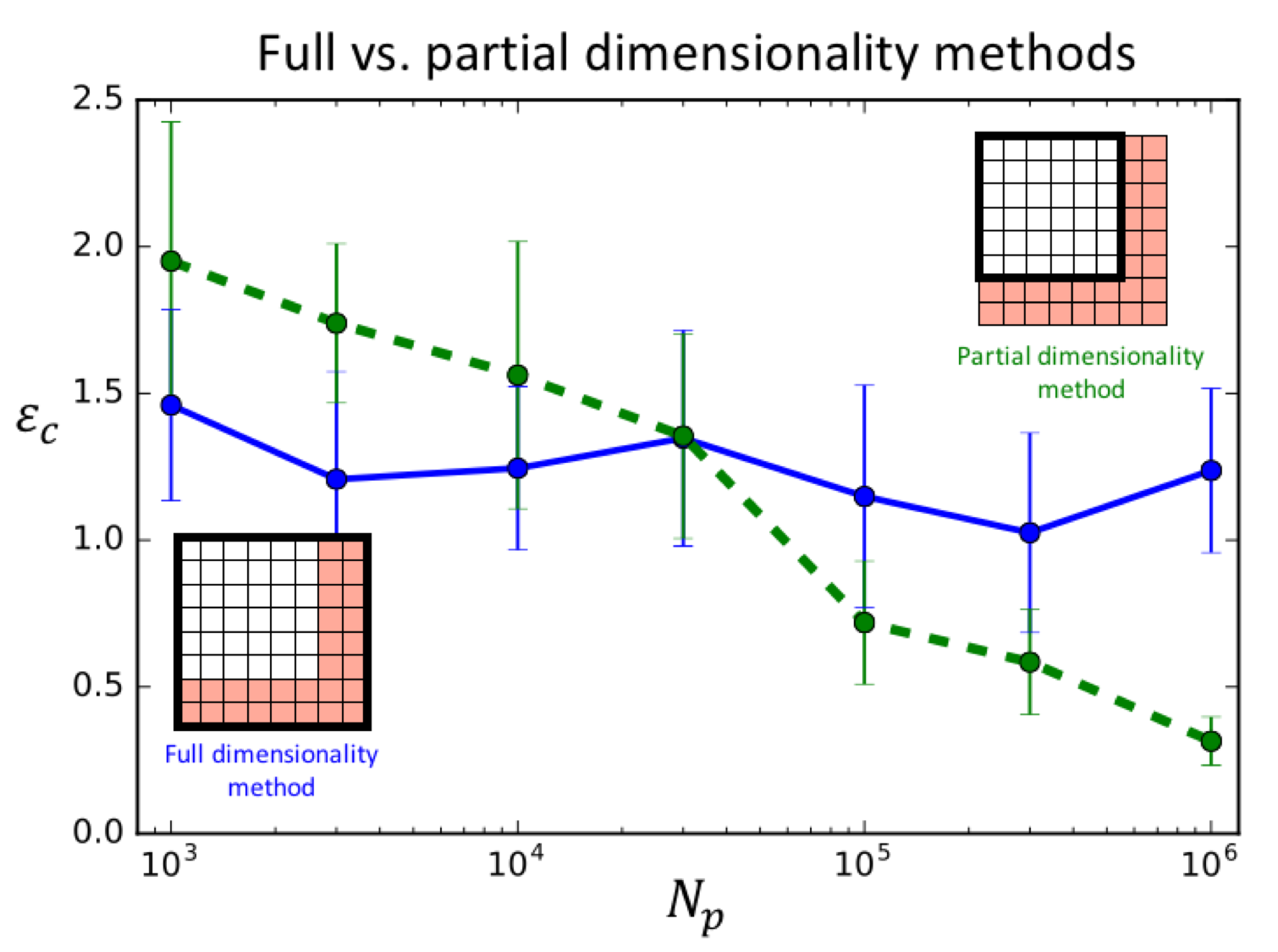}
\caption{Reconstruction error for complete dimensions, $\varepsilon_c$, vs. $N_p$. For a small number of partial samples, it is advantageous to perform the inference using all dimensions of the data (full dimensionality method - blue), whereas when there is a large amount of partial data it is best to perform the tensor decomposition on the lower dimensionality problem (partial dimensionality method - green). $N_f = 1e4$. The schematics illustrate both methods, with the bold box indicating the elements of the second order tensor being analyzed, and red boxes represent elements associated with the incomplete dimensions. Each data point is an average over 20 runs.}
\label{crossover_example}
\end{figure}

We consider two options for recovering the parameters corresponding to the complete dimensions, $\textbf{A}_c$.
The first, which we call the \emph{full dimensionality method},
makes use of all the data available. That is, if certain dimensions
of an observation $\textbf{x}_n$ are missing, we still use
the non-missing dimensions to assist in estimating the tensors
$\hat{\bf{S}}$ and $\hat{\bf{T}}$. (See the schematic insets in
Figure~\ref{crossover_example} to see what elements of the second order
tensor we take into account when performing the decomposition). Once we have learned the topics for all dimensions, we can observe only the complete dimensions which are of interest.

The full dimensionality method results in decomposing tensors in
which each element may have a different uncertainty. In contrast, the
second approach---which we call the \emph{partial dimensionality
  method}---discards any dimensions that are not always present in
the data and performs the inference on the lower dimensional problem
consisting of only the elements which are always observed.  Tensor
elements in the partial dimensionality method will have equal
uncertainties, but will include fewer correlations.

In Figure~\ref{features_leading_to_concept}, we provide some intuition
for the factors affecting the performance of the two methods.  In
Figure~\ref{features_leading_to_concept}~(a) we illustrate the effect of using only a subset of the dimensions for inference. Our goal is to infer the topic structure of six out of the ten dimensions in the data. To perform such inference, we must use the dimensions whose structure we try to infer, but may ignore some of the other dimensions. Figure~\ref{features_leading_to_concept} (a) demonstrates that for a given number of samples, ignoring the extra dimensions leads to larger inference error, as important information about the structure of the data is ignored.

To contrast the simple case where no data is missing, in Figure~\ref{features_leading_to_concept} (b), we demonstrate the effect of partial data on the topics recovery error. We present the reconstruction error using the full dimensionality method where for $N_f$ full samples all the dimensions in the data are observed, and for $N_p$ partial samples only 6 of the 10 dimensions are observed in the data. These partial data allow us to better estimate the moments associated only with
those 6 dimensions, but the error in the remaining moments is not
reduced by these additional observations. The relatively flat lines
suggest that the recovery error is dominated by the moments
with the highest uncertainty---something that the partial data cannot
fix---and therefore collecting additional partial data does little to improve the inference of the full dimensionality method.

Finally, in Figure~\ref{crossover_example}, we plot the reconstruction
error as a function of $N_p$ at a constant $N_f$ for both the full
dimensionality and the partial dimensionality methods.  As seen in
Figure~\ref{features_leading_to_concept} (b), additional partial
samples do not help the full dimensionality method, which is
dominated by the least accurate tensor elements.  In contrast, with
the partial dimensionality method, the topics reconstruction error
steadily decreases, as the estimates $\hat{\bf{S}}$ and $\hat{\bf{T}}$
converge to their true value (and as long as that subset of the true
topics matrix $\textbf{A}$ associated with those dimensions satisfies
the incoherence requirements in \citet{anandkumar2014tensor},
$\hat{\textbf{A}}_c$ will converge to $\textbf{A}_c$).  However, for
small amounts of partial data, because we are effectively working on
data with lower dimensionality, our reconstruction error is worse
compared to the results of the full dimensionality model---or put
another way, when all the tensor elements have similar uncertainties,
it makes sense to use all the correlations that we have available.
The existence of this cross-over suggests an opportunity for an
algorithm that can attain the better recovery in all cases.  We derive
such an algorithm in Section~\ref{sec:WTPM}.

\section{The Weighted Tensor Power Method (WTPM)}
\label{sec:WTPM}

In Section~\ref{sec:motivation}, we demonstrated that the choice of
whether to use the dimensions with missing data depends on the
\emph{relative} quality of the moment estimates.  In section \ref{sec:motivation} we gave an intuition for why if the
dimensions with missing data did not have \emph{significantly} less
data than the dimensions with complete data, it would make sense to
include those moments in order to get a more complete correlation
structure of the data.  However, since the quality of the inference is
dominated by the noisiest tensor elements, it does not make sense to
include tensor elements that may be of significantly lower
quality---as would be the case if the dimensions with missing data had
significantly less data associated with them than the complete
dimensions.  

In this section we introduce the \emph{Weighted Tensor Power Method} (WTPM)
for spectral learning with missing data.  It is more general than the
motivating example in Section~\ref{sec:motivation} in that we
allow different dimensions to have different levels of missingness,
but we shall also need to require that each dimension is missing at
random. (This assumption is required for our derivation. We will show that our approach is robust
to other missingness structures in Section~\ref{sec:synthetic}).  The key idea
behind our approach is the intuition that we would like to use all the
moments of the data, but assign higher importance to moments for which
we have smaller uncertainties. A natural and intuitive way to perform
this task is by performing tensor decomposition on weighted tensors. The weighting consists of multiplying every element in the empirical tensor estimates,
$\hat{\bf{S}}$ and $\hat{\bf{T}}$, by some weight which represents the
importance we give that specific element. Thus, the elements of the weighted tensors, $\hat{\bf{S}}^*$ and $\hat{\bf{T}}^*$, are given by
\begin{align}
\hat{S}_{d_1d_2}^* & = w_{d_1d_2} \hat{S}_{d_1d_2} \label{eq:weighted_S} \\
\hat{T}_{d_1d_2d_3}^* &= w_{d_1d_2d_3} \hat{T}_{d_1d_2d_3} \label{eq:weighted_T}.
\end{align}
To apply the efficient algorithms
of standard tensor decomposition methods, the weighting scheme must
also maintain the structure of the tensors as sums of $K$ rank 1
tensors, similar to moment equations (\ref{eq:S_decomposition})
and (\ref{eq:T_decomposition}). To maintain such a structure, we write
the weights as a product of the weights associated with every
dimension contributing to the tensor:
\begin{align}
\begin{split}
w_{d_1d_2} & = w_{d_1}w_{d_2} \\
w_{d_1d_2d_3} & = w_{d_1}w_{d_2}w_{d_3}.
\end{split}
\label{eq:weighting_as_rescaling}
\end{align}
This weighting scheme allows us to rewrite the moments as
\begin{align*}
S_{d_1d_2}^* & = w_{d_1d_2} S_{d_1d_2} = \sum_k s_k w_{d_1d_2} a_{d_1k} a_{d_2k}\\
& = \sum_k s_k (w_{d_1} a_{d_1k}) (w_{d_2} a_{d_2k}) = \sum_k s_ka_{d_1k}^* a_{d_2k}^*.
\end{align*}
where we defined $a_{dk}^*=a_{dk}w_d$. An analogous expression can be written for the third-order tensor $T^*$. Written in this form, it is clear that the weighted tensors also have the structure of a sum of $K$ rank 1
tensors, and therefore can be decomposed using standard tensor decomposition methods.

This form of the weights enjoys the additional advantage of allowing
us to view the weighting of the moments tensors as a rescaling of the
dimensions in the topics and data, $x_{dn}^*=x_{dn}w_d$ and $a_{dk}^*=a_{dk}w_d$, providing an intuitive explanation for the effect of weighting the moments. This view also allows us to perform the weighting on the data itself, multiplying each dimension by $w_i$, rather than on the moments.

Finally, once we perform the decomposition and learn the rescaled
topic matrix, we can rescale this matrix back by dividing its $d^{th}$ row by
$w_d$ to reconstruct the \emph{entire} topic matrix, and not just the
part corresponding to the complete dimensions. However, this full
reconstruction is very susceptible to noise as error in reconstructing
the incomplete dimensions will be amplified when dividing by the small
weights associated with them.

\paragraph{Running Time}

A consequence of viewing the weighting as a rescaling of the data is that the WTPM has the same running time as a standard tensor decomposition
method, as it can be implemented by scaling each dimension of the data
by $w_d$ and then applying one's tensor decomposition method of
choice.

\paragraph{Choice of Weights} 
The question remains of what weights $w_d$ to assign to each
dimension. Our approach is to choose weights that minimize the inference error. In this way, optimal weights can be computed for any model for which one can derive expressions for the inference error as a function of moment estimations errors and any other parameters affected by the weighting.

Demonstrating our approach, we first present the choice of optimal weights for the Gamma-Poisson model. \citet{podosinnikova2015rethinking} drew and expanded on
sample complexity bounds results from \citet{anandkumar2012spectral}
to show that the inference error has two main contributions, one which
scales as $\mathbb{E}\left[||\hat{\textbf{S}}-\textbf{S}||_F\right]/L^2$, and one which scales as
$\mathbb{E}\left[||\hat{\textbf{T}}-\textbf{T}||_F\right]/L^3$. By choosing our
weights to minimize these two ratios we can derive the weights which
minimize the topics reconstruction error:

\begin{lemma}
Assume that the magnitude of all elements in $\bf{A}$ are comparable, and that the probabilities of elements in the data to be missing are independent of each other. For high dimensional problems, an intuitive upper bound $\phi(\mathbf{w})$, of the expected topic reconstruction error, $E_{\mathrm{infer}}$, under the GP model is locally minimized by setting:
\begin{align*}
w_d \propto p_d,
\end{align*}
where $p_d$ is the probability of an element in the $d^{th}$ dimension to be observed in the data. 
\label{lem:optimal_weights}
\proof 
The upper bound $\phi(\mathbf{w})$ on $E_{\mathrm{infer}}$ is essentially the sum of the Jensen gaps of the expected sampling errors, $\mathbb{E}\left[||\hat{\textbf{S}}-\textbf{S}||_F\right]$ and $\mathbb{E}\left[||\hat{\textbf{T}}-\textbf{T}||_F\right]$.

See Appendix~\ref{appendix:optimal_weights} for details.
\end{lemma}

An immediate corollary of Lemma \ref{lem:optimal_weights} is that weighting the dimensions according to degrees of missingness is provably better for inference than the unweighted model. In Appendix~\ref{appendix:optimal_weights}, we show that under stronger assumptions, where we neglect the structure in the data, we can obtain globally optimal weighting by the choice $w_d \propto \sqrt{p_d}$. But in experimentation, we find the performance of both to be indistinguishable.

We note that  in our actual WTPM we choose not to rescale the complete dimensions, thus we simply set the proportionality constant to 1 and use the scaling $w_d = p_d$. In practice, we find that choosing the weights according to Lemma \ref{lem:optimal_weights} yields good results even when there is a large variance in the magnitude of the elements of $\bf{A}$. 

Additionally, we find that the weighting proposed in Lemma \ref{lem:optimal_weights} leads to good inference results for most models, including Gaussian mixtures. Thus, in many cases, one may forgo derivations involving analytical expressions for the inference error in favor of the the heuristic of choosing $w_d = p_d$. For example, while inference error bounds (in terms of the weights) exist for Gaussian Mixtures, the necessary conditions for these bounds to hold render optimization intractable. 



Finally, we note that the mixture of Gaussians also requires an
estimation of the variance, and reweighing dimensions will change a
spherical Gaussian into an elliptical one.  We provide details on how the weighting affects the variance estimation process in appendix~\ref{appendix:elliptical_gaussians}.

\section{Results}

In this section we compare our WTPM with the full and partial
dimensionality methods described in Section~\ref{sec:motivation}. We
test our method on synthetic, semi-synthetic, and real data sets for
different structures of missing data and show that in all cases the
WTPM interpolates between the more naive methods, always performing as
well as or better than the best of them.

\subsection{Synthetic Examples}
\label{sec:synthetic}

\begin{figure*}[t]
  \centering
  \begin{subfigure}{.4\linewidth}
    \centering
    \includegraphics[width=\linewidth]{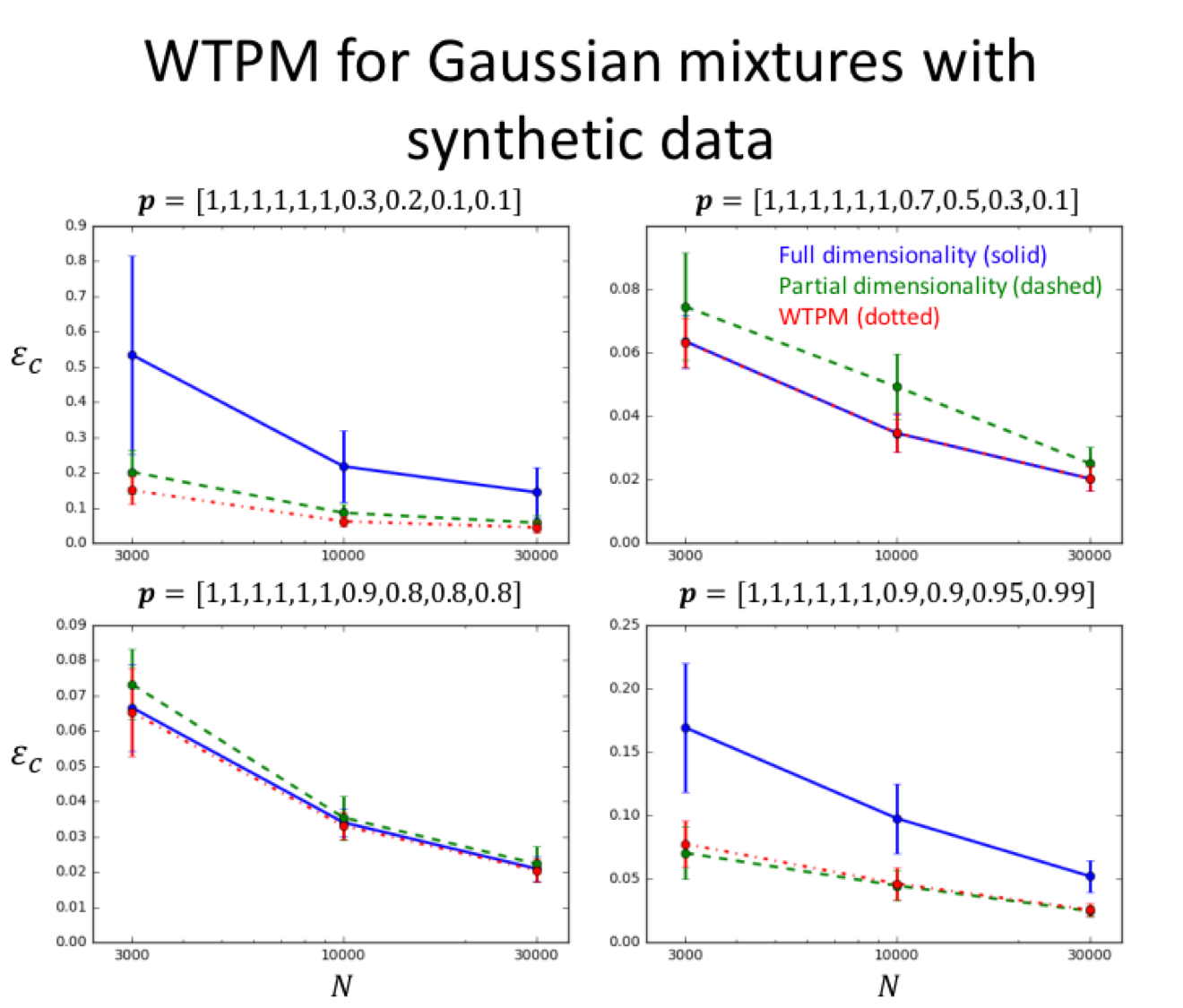}
  \end{subfigure}
  \rulesep
    \begin{subfigure}{.4\linewidth}
    \centering
  \includegraphics[width=\linewidth]{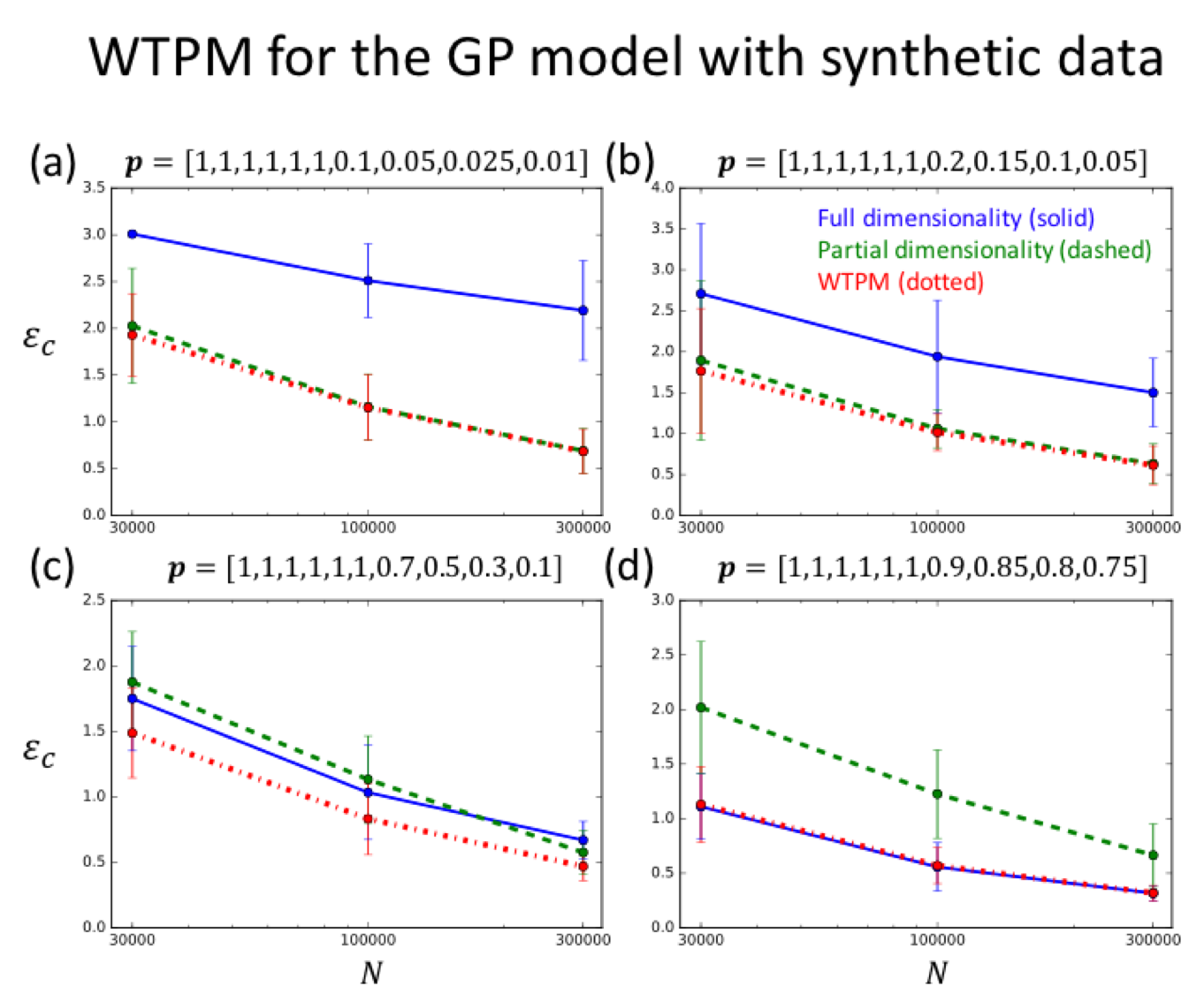}
    \end{subfigure}
    \caption{Reconstruction error for the complete dimensions of b
      synthetic data, $\varepsilon_c$, vs. $N$, where every incomplete
      dimension has a different probability, $p_d$, to be observed for
      a mixture of Gaussians model (left) and gamma-Poisson model
      (right). Each data point is an average over 20 runs.  Whether
      the partial or full dimensionality method does better depends on
      the missingness rate; our WTPM \emph{always} performs as well as
      the better of the baselines.}
    \label{fig:random_missing_data}
\end{figure*}

\paragraph{Gaussian Mixtures}
Figure \ref{fig:random_missing_data} (left) compares the WTPM with the full and partial dimensionality methods for synthetic data of Gaussian mixtures with $D=10$, $K=4$ and $4$ missing dimensions. The inference error for the complete dimensions, $\varepsilon_c$ is plotted vs. the number of samples, $N$. Each sub-figure shows the results for a different missingness pattern, given as a vector, $\bf{p}$, representing the probability of each dimension to be present in the data. The first six dimensions in each instance are complete dimensions and are always observed ($p=1$), and the last four incomplete dimensions are present in the data with different probabilities. The centers of all Gaussians were sampled from a normal distribution $A_{ij}\sim\mathcal{N}(0,100)$ for all $(i,j)$, and the variance of the spherical Gaussians is $\sigma^2=100$. $\pmb{\pi}$ was sampled from a Dirichlet distribution, $\pmb{\pi} \sim \text{Dir}(\bf{1})$. In all cases, the WTPM always did as well as or better than both the full and partial dimensionality method. We emphasize that because in some instances the partial dimensionality method outperforms the full dimensionality method and in some instances the opposite is true, always matching the best of the two methods with one algorithm is already advantageous, even when the WTPM does not outperform the two other methods.

\paragraph{Gamma-Poisson}
Next, we investigate the performance of our WTPM and compare it to the performance of the full and partial dimensionality methods for data generated from the topics presented in Figure~\ref{features_leading_to_concept}. In Figure~\ref{fig:random_missing_data} (right) we present results for experiments on four different choices of missingness patterns, given by the vector $\textbf{p}$. When the incomplete dimensions are missing with high probabilities (small $p$ for the incomplete elements, Figure~\ref{fig:random_missing_data} (a-b)), the moments including the incomplete dimensions are very noisy and including them in the inference by using the full dimensionality method (blue) leads to a higher reconstruction error compared with the partial dimensionality method (green). The opposite is true when the incomplete dimensions are missing with low probability (large $p$ for the incomplete elements, Figure~\ref{fig:random_missing_data} (d)) and the full dimensionality method performs better by using the additional correlations which are less noisy in this case.

As with the Gaussian mixtures, the WTPM (red) does as well as or
better than the best of the two baselines in all cases. In cases where
values of $p$ for the incomplete dimension are distributed over a wide
range of values (Figure~\ref{fig:random_missing_data} (c)), the WTPM
significantly outperforms both other methods, as it optimizes the
amount of information in every moment, while properly down-weighting
each moment according to its uncertainty.

\begin{figure*}[t]
\centering
\begin{subfigure}{.3\linewidth}
\centering
\includegraphics[width=\linewidth]{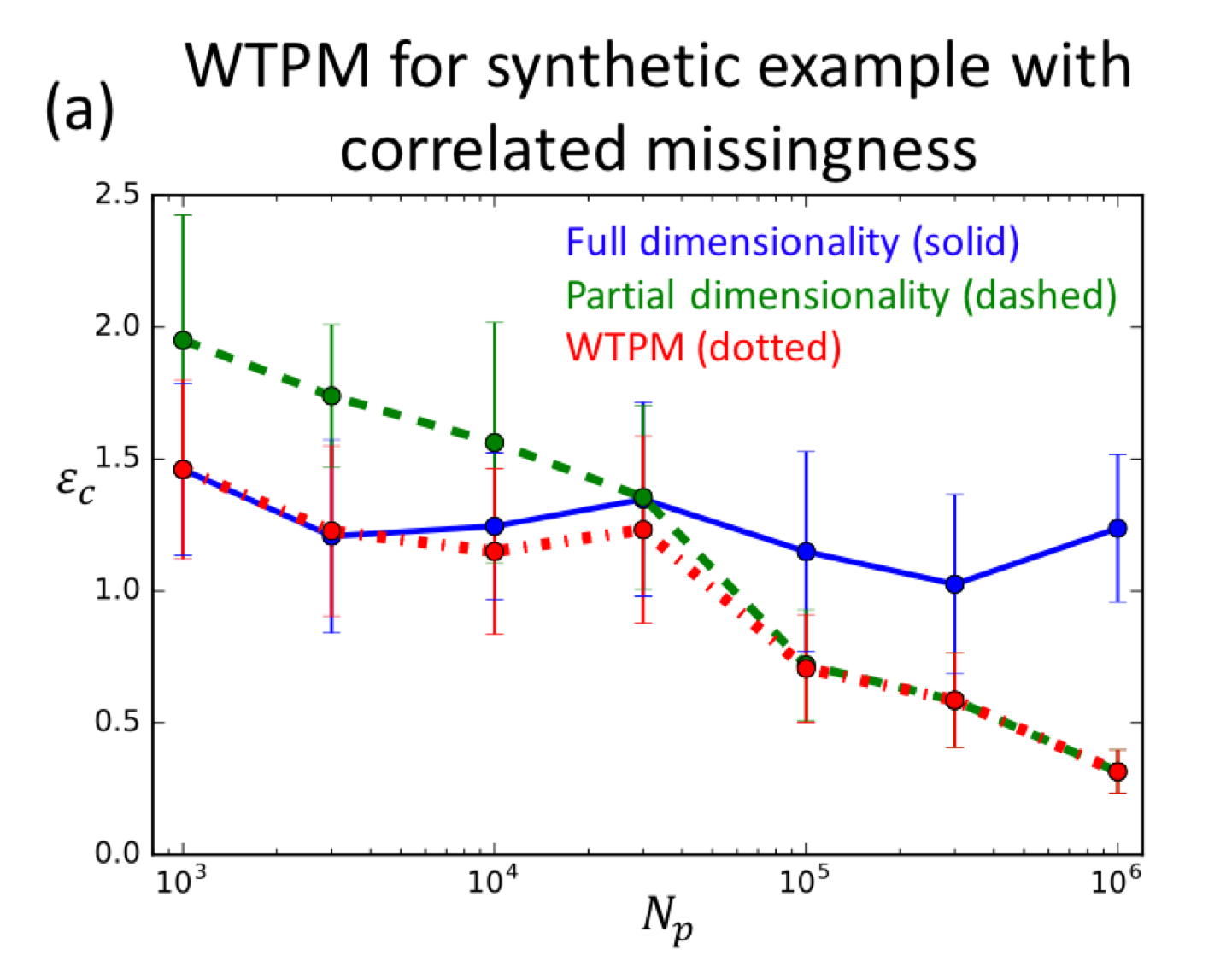}
\end{subfigure}
\rulesep
\begin{subfigure}{.6\linewidth}
\centering
\includegraphics[width=\linewidth]{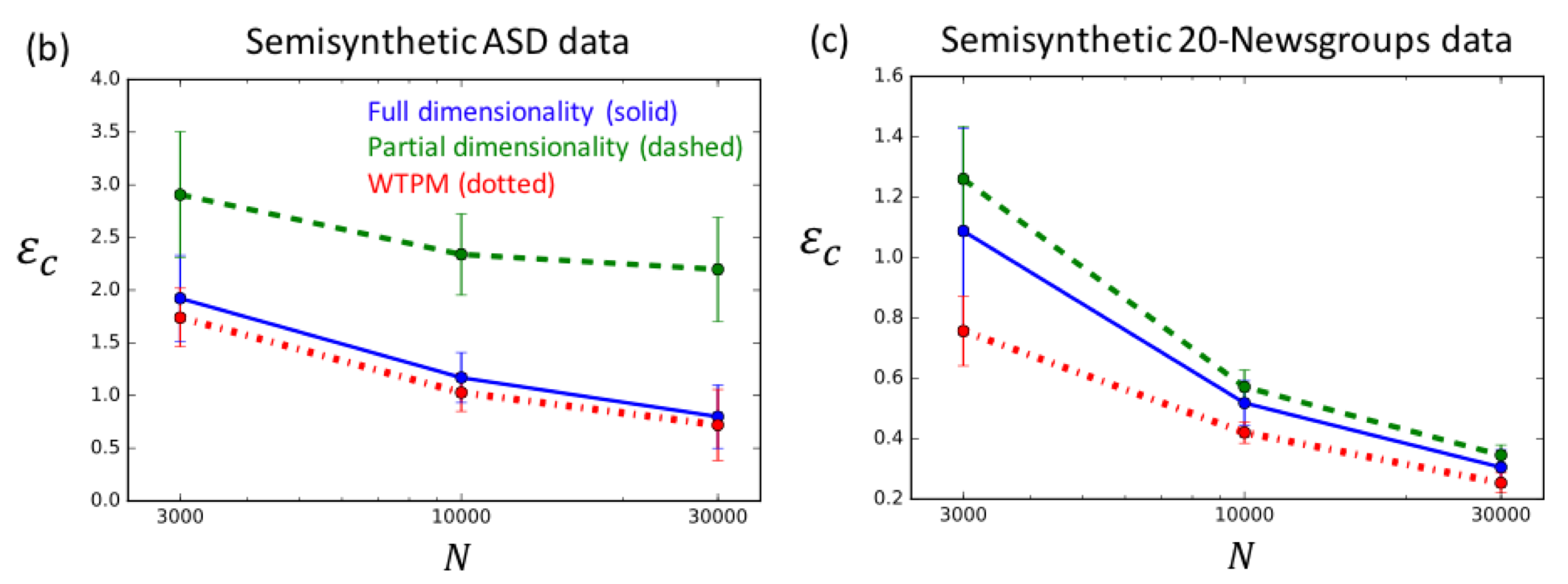}
\end{subfigure}
\caption{(a) Comparison of the WTPM with the full and partial
  dimensionality methods for synthetic topics with correlated
  missingness. Despite the violation of the missingness independence
  assumption, the WTPM still performs as well as the better of the two
  baselines. (b-c) Reconstruction error for complete dimensions of
  semi-synthetic data, $\varepsilon_c$, vs. $N$, in cases where every
  incomplete dimension has a different probability, $p_d$, to be
  observed. $p_d$ is sampled uniformly from $[0,0.9]$ for 64 out of
  the 128 dimensions in the ASD data (left) and for 116 out of the 175
  dimensions in the 20-Newsgroups data (right).}
\label{fig:random_missing_data_semisynthetic}
\end{figure*}

Finally, in Figure~\ref{fig:random_missing_data_semisynthetic}(a), we
present results based on the missingness pattern in
Figure~\ref{data_structure}(b) and the synthetic model in
Figure~\ref{crossover_example}, where all partial dimensions are
either present or missing together.  This missingness violates the
assumptions required in Lemma~\ref{lem:optimal_weights}, the WTPM is
robust to this violation and outperforms both full and partial
dimensionality methods in this case as well.

\subsection{Semi-synthetic Examples: 20 Newsgroups and Autism Comorbidities}
To investigate the effect of incomplete dimensions within the kinds of
sparse topic structure found in real data, we adopt a similar approach
to \citet{arora2013practical} and run still carefully-controlled
experiments on semi-synthetic data.  We use two real data sets to
create semi-synthetic data for the WTPM: word counts from 4 categories
of the 20-Newsgroups collection of news documents
\citep{TwentyNewsgroups}, filtered for words which appear in at least
$5\%$ of the documents ($D=175$, $K=4$, original $N=3,652$); and
counts of 64 common diagnoses from a collection electronic
health records of children with autism at ages 5 and 10
\citep{doshi2014comorbidity} ($D=128$, $K=5$, original $N=5,475$).
%

To create the semi-synthetic data from these data, we first run a
standard tensor decomposition on a real dataset to learn the
parameters of a GP model ($\textbf{D}$, $\textbf{c}$ and $b$), mimicking
the sparse data structure common to real world applications.  To
create the missing data structure, we first chose a set of missing
dimensions for each data set. For the ASD data we chose the 64
dimensions corresponding to observations of symptoms at age 5. For the
20-Newsgroups data we randomly chose 116 out of the 175
dimensions. For each one of the incomplete dimensions, $p_d$ was
sampled uniformly between 0 to 0.9, resulting in a complicated
structure of incomplete dimensions, with some dimensions missing in an
insignificant amount of data, and some dimensions nearly completely
unobserved.  We emphasize that while the original data sets used to
learn the topics are small, the semi-synthetic data set is large,
allowing us to test the performance of our algorithm for a large
amount of data.

Figure~\ref{fig:random_missing_data_semisynthetic} compares the WTPM
with the full and partial dimensionality methods for the
semi-synthetic data and a random structure of missingness in the data.
The WTPM performs as well or better than both the full and partial
dimensionality method across the entire wide range of missingness
patterns tested.

\subsection{Real Data Example: Sepsis ICU data}

As a final demonstration of the WTPM, we use the method to learn the structure of data on sepsis patients collected in ICUs. The data is obtained from the Medical Information Mart for Intensive Care III (MIMIC-III) database \citep{johnson2017mimic}. The database contains more than 18,000 patients, and every patient's vitals were sampled on average at 14 different times, resulting in about 250,000 samples. Each sample is represented as a 47 dimensional vector containing information on a patient such as the vitals, lab results, demographics, etc. We wish to learn the structure of the data by modeling it as a mixture of Gaussians. The data collected in the ICU is often incomplete, and the missingness rate for different dimensions ranges from 0 to 0.75, making it an ideal dataset to test with our WTPM. We define as complete dimensions all the dimensions whose missingness rate is smaller than 0.05, resulting in 24 complete dimensions. We learn a mixture of Gaussian model for the complete dimensions with $K=6$ topics using the full and partial dimensionality methods, as well the WTPM.

\begin{figure}[t]
\centering
\includegraphics[width=0.5\linewidth]{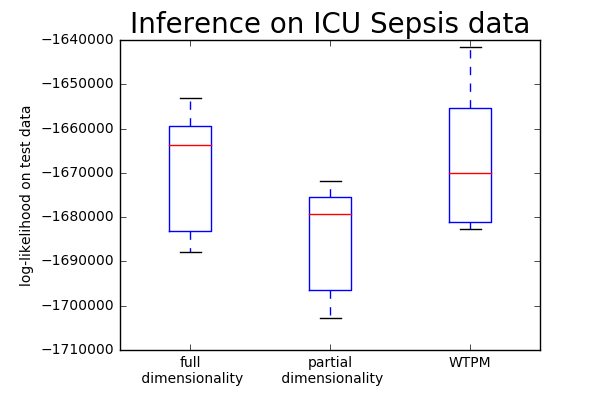}
\caption{Log-likelihood of held-out test data for the full and partial dimensionality methods, as well as the WTPM. The box plot represents results from 20 different partitioning of the data in training and test data sets, where 0.2 of the data was held out. The WTPM performs as well as the full dimensionality method and outperforms the partial dimensionality method.}
\label{fig:real_data}
\end{figure}

In Figure \ref{fig:real_data} we demonstrate that the WTPM performs as well as the full dimensionality method, which performs significantly better than the partial dimensionality method.  The WTPM performs well despite the noisiness of the data: the missingness rate spans a wide range of values across different dimensions and is non-zero even for most of the dimensions we define as complete. Furthermore, given the nature of data collection in the ICU, we believe the missingness pattern is likely to be strongly correlated; the WTPM is robust to these assumption violations.

\section{Discussion}

The derivation of the ideal weights presented in Lemma
\ref{lem:optimal_weights} assumes the missingness of different
elements is uncorrelated, and also ignores the structure of the topics
matrix. However, we have demonstrated that the WTPM is robust to
violations of these assumptions:
Figure~\ref{fig:random_missing_data_semisynthetic} demonstrates that
the WTPM method works with a perfectly correlated missingness pattern
(an element is missing in a sample if and only if all other incomplete
dimensions in the sample are missing). The effectiveness of the WTPM
on the semi-synthetic data generated from sparse topics matrices with
significant magnitude differences between their non-zero elements
demonstrates the insensitivity to the structure of the topics
matrix. In future work, it would be interesting to check whether the
effectiveness of the WTPM can be further improved by taking into
account the structure of the topics and correlations in the
missingness. For example, taking into account the structure of the
data could mean that it would be advantageous to weight differently
dimensions which on average have higher values compared with other
dimensions.

While we focused on Gaussian mixtures and the gamma-Poisson model in this work, the core idea
behind the WTPM can be applied to method-of-moment-based inference for
other models as well. The form of the sample complexity bounds for LDA 
\citep{anandkumar2012spectral}, HMM \citep{hsu2012spectral} and other topic models such as correlated topic models and Pachinko allocation \citep{arora2013practical,blei2007correlated,li2006pachinko} are quite similar to the bounds calculated by \citet{podosinnikova2015rethinking}, and we therefore expect the application of the WTPM to these models to only require a derivation similar to that presented in Appendix~\ref{appendix:optimal_weights}.

Another interesting direction is asking what we can say about the
parameters of the dimensions that are incompletely observed.  In this
work, we demonstrated that lower-quality moment estimates from the
incomplete dimensions could assist with the recovery of the parameters
associated with the complete dimensions.  The WTPM approach does give
us values for all of the parameters---including ones associated with
incomplete dimensions.  These parameters can be straightforwardly
up-weighted by $\frac{1}{w_d}$ to get parameter-values for the original
data scaling.  However, if those parameters are poorly
estimated---which is likely, if their associated moment estimates were
poor---this rescaling will not result in smaller errors compared with a standard
tensor decomposition approach.  It is an open question if it is
possible to get better recovery of these parameters as well. 

Finally, we note that given an implementation of the tensor power
method described in \citet{anandkumar2014tensor}, performing the WTPM
is very straightforward, and requires only multiplying the empirical
estimates $\hat{\textbf{S}}$ and $\hat{\textbf{T}}$ by
$\textbf{p}\otimes\textbf{p}$ and
$\textbf{p}\otimes\textbf{p}\otimes\textbf{p}$ respectively, before
performing the decomposition. In practice, $\hat{\textbf{T}}$ is
frequently not calculated explicitly in order to avoid a cubic
dependence of the runtime on the dimensionality. Rather, it is
possible to use an implicit calculation of $\hat{\textbf{T}}$ that
reduces the runtime to $O(DK+nnz(\textbf{X}))$, where
$nnz(\textbf{X})$ is the number of non-zero elements in $\textbf{X}$
\citep{zou2013contrastive}. However, the method proposed in
\citet{zou2013contrastive}, as well as other methods which do not
explicitly calculate $\hat{\textbf{T}}$ can only work when there are no
missing elements in the data. While the explicit calculation of the third moments does not scale well with dimensionality of the data, its runtime is independent of document length once the data is converted into word-count format, and should therefore out-perform inference methods such as Gibbs-sampling for data consisting of very long documents.

\section{Related Work}
Tensor decomposition has been studied for nearly a century, but has
recently gained popularity within many different scientific
communities \citep{kolda2009tensor}.  In particular, a series of works
have shown that tensor decomposition methods can recover the
parameters of many classes of latent variable models
\citep{anandkumar2014tensor,arora2012learning,collins2012tensor,zou2013contrastive,parikh2011spectral,podosinnikova2015rethinking}.
These works generally provide connections between the quality of the
moment estimates and the model parameter estimates as a step toward
deriving high-probability bounds on the quality of model parameter
estimates; however, none of them address the question of what to do
when moment estimates are of only moderate and unequal quality.

More broadly, it is well-known that decompositions of empirical
moments may result in model parameter estimates that are far from the
maximum-likelihood parameter values, especially in the presence of
model mismatch. \citet{zhang2014spectral} and
\citet{shaban2015learning} both show that in certain circumstances,
using the spectral parameter estimates as initialization for an
optimization procedure can result in the discovery of the true
maximum-likelihood parameter settings.  \citet{tran2016spectral} use
an M-estimation framework to regularize the parameter estimates.  In
the specific case of HMMs, \citet{huang2013spectral} show how to
combine cross-sectional and sequence data to produce better parameter
estimates.  \citet{cheng2015model} investigate the effect
uncertainties in the second order tensor elements have on its spectral
structure in order to bound the numbers of topics.  Again, none of
these works address inference in the presence of unequal quality
moment estimates.

Finally, there exists a large literature relating to algorithms for
tensor decompositions.  \citet{jain2014provable} discuss decompositions
when tensor elements are missing, distinct from our case in which
certain dimensions are not always recorded.  Weighted matrix
factorization methods such as \citet{srebro2003weighted}, are closely
related in spirit to the algorithm we presented and perform a weighted
decomposition of higher order tensors.

\section{Conclusion}

In this work, we derived and presented the Weighted Tensor
Decomposition Method (WTDM) for method-of-moment inference in latent
variables with missing data.  In situations where some dimensions are
incompletely observed, we show that we can still leverage the
associated low quality moment estimates to get better parameter
estimation than simply ignoring that information, while minimizing
inference errors due to moments that are too noisy. Our method is easy to
implement and enjoys the same large-sample recovery guarantees as
standard tensor decomposition-based inference methods; we demonstrate
its benefits for two models and our derivations are general to other
models as well.

\bibliographystyle{plainnat}
\bibliography{WTPM_arxiv}

\begin{thebibliography}{26}
\providecommand{\natexlab}[1]{#1}
\providecommand{\url}[1]{\texttt{#1}}
\expandafter\ifx\csname urlstyle\endcsname\relax
  \providecommand{\doi}[1]{doi: #1}\else
  \providecommand{\doi}{doi: \begingroup \urlstyle{rm}\Url}\fi

\bibitem[Anandkumar et~al.(2012)Anandkumar, Foster, Hsu, Kakade, and
  Liu]{anandkumar2012spectral}
Anima Anandkumar, Dean~P Foster, Daniel~J Hsu, Sham~M Kakade, and Yi-Kai Liu.
\newblock A spectral algorithm for latent dirichlet allocation.
\newblock In \emph{Advances in Neural Information Processing Systems}, pages
  917--925, 2012.

\bibitem[Anandkumar et~al.(2014)Anandkumar, Ge, Hsu, Kakade, and
  Telgarsky]{anandkumar2014tensor}
Animashree Anandkumar, Rong Ge, Daniel~J Hsu, Sham~M Kakade, and Matus
  Telgarsky.
\newblock Tensor decompositions for learning latent variable models.
\newblock \emph{Journal of Machine Learning Research}, 15\penalty0
  (1):\penalty0 2773--2832, 2014.

\bibitem[Arora et~al.(2012)Arora, Ge, and Moitra]{arora2012learning}
Sanjeev Arora, Rong Ge, and Ankur Moitra.
\newblock Learning topic models--going beyond svd.
\newblock In \emph{Foundations of Computer Science (FOCS), 2012 IEEE 53rd
  Annual Symposium on}, pages 1--10. IEEE, 2012.

\bibitem[Arora et~al.(2013)Arora, Ge, Halpern, Mimno, Moitra, Sontag, Wu, and
  Zhu]{arora2013practical}
Sanjeev Arora, Rong Ge, Yonatan Halpern, David~M Mimno, Ankur Moitra, David
  Sontag, Yichen Wu, and Michael Zhu.
\newblock A practical algorithm for topic modeling with provable guarantees.
\newblock In \emph{ICML (2)}, pages 280--288, 2013.

\bibitem[Blei and Lafferty(2007)]{blei2007correlated}
David~M Blei and John~D Lafferty.
\newblock A correlated topic model of science.
\newblock \emph{The Annals of Applied Statistics}, pages 17--35, 2007.

\bibitem[Blei et~al.(2003)Blei, Ng, and Jordan]{blei2003latent}
David~M Blei, Andrew~Y Ng, and Michael~I Jordan.
\newblock Latent dirichlet allocation.
\newblock \emph{Journal of machine Learning research}, 3\penalty0
  (Jan):\penalty0 993--1022, 2003.

\bibitem[Cardoso and Souloumiac(1993)]{cardoso1993blind}
Jean-Fran{\c{c}}ois Cardoso and Antoine Souloumiac.
\newblock Blind beamforming for non-gaussian signals.
\newblock In \emph{IEE Proceedings F (Radar and Signal Processing)}, volume
  140, pages 362--370. IET, 1993.

\bibitem[Cheng et~al.(2015)Cheng, He, and Liu]{cheng2015model}
Dehua Cheng, Xinran He, and Yan Liu.
\newblock Model selection for topic models via spectral decomposition.
\newblock In \emph{AISTATS}, 2015.

\bibitem[Collins and Cohen(2012)]{collins2012tensor}
Michael Collins and Shay~B Cohen.
\newblock Tensor decomposition for fast parsing with latent-variable pcfgs.
\newblock In \emph{Advances in Neural Information Processing Systems}, pages
  2519--2527, 2012.

\bibitem[Doshi-Velez et~al.(2014)Doshi-Velez, Ge, and
  Kohane]{doshi2014comorbidity}
Finale Doshi-Velez, Yaorong Ge, and Isaac Kohane.
\newblock Comorbidity clusters in autism spectrum disorders: an electronic
  health record time-series analysis.
\newblock \emph{Pediatrics}, 133\penalty0 (1):\penalty0 e54--e63, 2014.

\bibitem[Hsu and Kakade(2013)]{hsu2013learning}
Daniel Hsu and Sham~M Kakade.
\newblock Learning mixtures of spherical gaussians: moment methods and spectral
  decompositions.
\newblock In \emph{Proceedings of the 4th conference on Innovations in
  Theoretical Computer Science}, pages 11--20. ACM, 2013.

\bibitem[Hsu et~al.(2012)Hsu, Kakade, and Zhang]{hsu2012spectral}
Daniel Hsu, Sham~M Kakade, and Tong Zhang.
\newblock A spectral algorithm for learning hidden markov models.
\newblock \emph{Journal of Computer and System Sciences}, 78\penalty0
  (5):\penalty0 1460--1480, 2012.

\bibitem[Huang and Schneider(2013)]{huang2013spectral}
Tzu-Kuo Huang and Jeff~G Schneider.
\newblock Spectral learning of hidden markov models from dynamic and static
  data.
\newblock In \emph{ICML (3)}, pages 630--638, 2013.

\bibitem[Hyv et~al.(1999)]{hyv1999fast}
Aapo Hyv et~al.
\newblock Fast and robust fixed-point algorithms for independent component
  analysis.
\newblock \emph{IEEE Transactions on Neural Networks}, 10\penalty0
  (3):\penalty0 626--634, 1999.

\bibitem[Jain and Oh(2014)]{jain2014provable}
Prateek Jain and Sewoong Oh.
\newblock Provable tensor factorization with missing data.
\newblock In \emph{Advances in Neural Information Processing Systems}, pages
  1431--1439, 2014.

\bibitem[Johnson et~al.(2017)Johnson, Stone, Celi, and
  Pollard]{johnson2017mimic}
Alistair~EW Johnson, David~J Stone, Leo~A Celi, and Tom~J Pollard.
\newblock The mimic code repository: enabling reproducibility in critical care
  research.
\newblock \emph{Journal of the American Medical Informatics Association}, page
  ocx084, 2017.

\bibitem[Kolda and Bader(2009)]{kolda2009tensor}
Tamara~G Kolda and Brett~W Bader.
\newblock Tensor decompositions and applications.
\newblock \emph{SIAM review}, 51\penalty0 (3):\penalty0 455--500, 2009.

\bibitem[Li and McCallum(2006)]{li2006pachinko}
Wei Li and Andrew McCallum.
\newblock Pachinko allocation: Dag-structured mixture models of topic
  correlations.
\newblock In \emph{Proceedings of the 23rd international conference on Machine
  learning}, pages 577--584. ACM, 2006.

\bibitem[Mitchell(1997)]{TwentyNewsgroups}
Tom Mitchell.
\newblock {UCI} machine learning repository: Twenty newsgroups, 1997.
\newblock URL \url{https://archive.ics.uci.edu/ml/ datasets/Twenty+Newsgroups}.

\bibitem[Parikh et~al.(2011)Parikh, Song, and Xing]{parikh2011spectral}
Ankur~P Parikh, Le~Song, and Eric~P Xing.
\newblock A spectral algorithm for latent tree graphical models.
\newblock In \emph{Proceedings of the 28th International Conference on Machine
  Learning}, pages 1065--1072, 2011.

\bibitem[Podosinnikova et~al.(2015)Podosinnikova, Bach, and
  Lacoste-Julien]{podosinnikova2015rethinking}
Anastasia Podosinnikova, Francis Bach, and Simon Lacoste-Julien.
\newblock Rethinking lda: moment matching for discrete ica.
\newblock In \emph{Advances in Neural Information Processing Systems}, pages
  514--522, 2015.

\bibitem[Shaban et~al.(2015)Shaban, Farajtabar, Xie, Song, and
  Boots]{shaban2015learning}
Amirreza Shaban, Mehrdad Farajtabar, Bo~Xie, Le~Song, and Byron Boots.
\newblock Learning latent variable models by improving spectral solutions with
  exterior point method.
\newblock In \emph{UAI}, pages 792--801, 2015.

\bibitem[Srebro et~al.(2003)Srebro, Jaakkola, et~al.]{srebro2003weighted}
Nathan Srebro, Tommi Jaakkola, et~al.
\newblock Weighted low-rank approximations.
\newblock In \emph{Icml}, volume~3, pages 720--727, 2003.

\bibitem[Tran et~al.(2016)Tran, Kim, and Doshi-Velez]{tran2016spectral}
Dustin Tran, Minjae Kim, and Finale Doshi-Velez.
\newblock Spectral m-estimation with applications to hidden markov models.
\newblock \emph{AISTATS}, 2016.

\bibitem[Zhang et~al.(2014)Zhang, Chen, Zhou, and Jordan]{zhang2014spectral}
Yuchen Zhang, Xi~Chen, Denny Zhou, and Michael~I Jordan.
\newblock Spectral methods meet em: A provably optimal algorithm for
  crowdsourcing.
\newblock In \emph{Advances in neural information processing systems}, pages
  1260--1268, 2014.

\bibitem[Zou et~al.(2013)Zou, Hsu, Parkes, and Adams]{zou2013contrastive}
James~Y Zou, Daniel~J Hsu, David~C Parkes, and Ryan~P Adams.
\newblock Contrastive learning using spectral methods.
\newblock In \emph{Advances in Neural Information Processing Systems}, pages
  2238--2246, 2013.

\end{thebibliography}

\clearpage

\begin{appendices}

\section{Generative models and empirical moments}
\label{appendix:empirical_moments}

\begin{figure}[ht]
\centering
\includegraphics[width=0.5\linewidth]{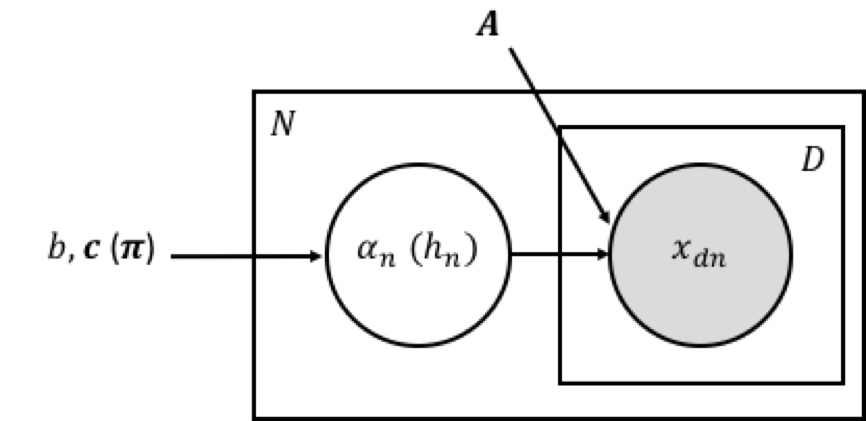}
\caption{Schematic illustration of the gamma-Poisson and mixture of Gaussians generative models.}
\label{fig:generative_model}
\end{figure}

\paragraph{Spherical Gaussian Mixtures}
The spherical Gaussian mixture model posits that the data matrix,
$\textbf{X}\in\mathbb{R}^{D\times N}$, consists of $N$ data points
represented as $D$ dimensional vectors. The generative process for the
$n^{th}$ data point, $\textbf{x}_n$, is
\begin{align*}
  h_n &\sim \text{Multinomial}(1,\pmb{\pi}), \\
  x_n|h_n,\textbf{A} &\sim \mathcal{N}(\textbf{a}_{h_n},\sigma^2).
\end{align*}

where $\textbf{a}_{h_n}$ is the $(h_n)^{th}$ column (topic) in the topics matrix $\textbf{A}\in \mathbb{R}^{D\times K}$, and $\pmb{\pi}\in \mathbb{R}^K$ represents the probability of data points to be drawn from each topic $(\sum_{k=1}^K \pi_k = 1)$. A schematic illustration is presented in
Figure~\ref{fig:generative_model}.

\citet{hsu2013learning} showed that if we estimate the variance of the Gaussians, $\sigma^2$, as the smallest eigenvalue of the covariance matrix, $ \mathbb{E}[\bf{x \otimes x}] - \mathbb{E}[\bf{x}] \otimes \mathbb{E}[\bf{x}]$, the empirical estimates

\begin{align}
\hat{\textbf{S}} &= \mathbb{E}[\bf{x} \otimes x] - \sigma^2I \\
\hat{\textbf{T}} &= \mathbb{E}[\textbf{x} \otimes \textbf{x} \otimes \textbf{x}] - \sigma^2 \sum_{i=1}^D(\mathbb{E}[\textbf{x}]\otimes \textbf{e}_i \otimes \textbf{e}_i \notag\\
  &+ \textbf{e}_i \otimes \mathbb{E}[\textbf{x}]\otimes \textbf{e}_i + \textbf{e}_i \otimes \textbf{e}_i \otimes \mathbb{E}[\textbf{x}]),
\label{eq:tensor_estimates_Gauss}
\end{align}

converge to the theoretical moments of the model -

\begin{align}
\textbf{S} & = \sum_k \pi_k \textbf{a}_k \textbf{a}_k^T, \label{eq:S_decomposition_Gauss}\\
\textbf{T} & = \sum_k \pi_k \textbf{a}_k \otimes \textbf{a}_k \otimes \textbf{a}_k.
\end{align}

\paragraph{Gamma-Poisson Model}
The second model we will focus on is the the gamma-Poisson (GP)
generative model described in \citet{podosinnikova2015rethinking}.
The GP model is closely related to latent Dirichlet allocation (LDA)
\citep{podosinnikova2015rethinking}.  In addition to its popular use
for modeling text corpora \citep{blei2003latent}, the GP model is also
relevant for capturing structure in applications where not all counts
may be recorded (such as what parts of the genome are sequenced in
genomics).  A schematic illustration is presented in
Figure~\ref{fig:generative_model}.

Formally, we represent the data as a matrix
$\textbf{X}\in\mathbb{N}_0^{D\times N}$ (that is, the data
$\textbf{X}$ is a matrix of non-negative integers), with every column
$\textbf{x}_n$ sampled according to
\begin{align*}
  \alpha_{nk} &\sim \text{Gamma}(c_k,b), \\
  x_d|\textbf{$\alpha$}_n,\textbf{A} &\sim \text{Poisson}([\textbf{A} \pmb{\alpha}_n]_d).
\end{align*}

Here, the global topics matrix $\textbf{A}\in \mathbb{R}_+^{D\times
  K}$ can be interpreted as the collection of rates of word $d$ in
topic $k$; following \citet{podosinnikova2015rethinking} and without
loss of generality we constrain the columns in $\textbf{A}$ to sum to
1. The observation-specific vector $\alpha_n\in\mathbb{R}^K$
determines the relative contribution of each of the $K$ topics for a
particular observation $n$. The parameters $b$ and
$\textbf{c}\in\mathbb{R}^K$ are constants that encode our prior about
both the length and relative popularities of the topics.  In the
context of text modeling, every element $x_{dn}$ can be thought of as
representing the number of times the $d^{th}$ word in the vocabulary
appears in the $n^{th}$ document, with the mean document length being
$L=\sum_kc_k/b.$

In this work, we will use the following second and third
order tensors, first introduced by \citet{podosinnikova2015rethinking}:
\begin{align}
\hat{\textbf{S}} & = \textrm{cov}(\textbf{x},\textbf{x}) - \textrm{diag}(\mathbb{E}(\textbf{x})) \\
\hat{T}_{d_1,d_2,d_3} & = \textrm{cum}(x_{d_1},x_{d_2},x_{d_3}) + 2\delta_{d_1d_2d_3}\mathbb{E}(x_{d_1})\notag\\
  & - \delta_{d_2d_3}\textrm{cov}(x_{d_1},x_{d_2}) - \delta_{d_1d_3}\textrm{cov}(x_{d_1},x_{d_2}) \notag\\ 
  & - \delta_{d_1d_2}\textrm{cov}(x_{d_1},x_{d_3})
\label{tensor_estimates}
\end{align}
where $\delta$ is the Kronecker delta and the second and third
cumulants are defined as
\begin{align*}
\textrm{cov}(x_{d_1},x_{d_2}) & = \mathbb{E}[(x_{d_1}-\mathbb{E}[x_{d_1}])(x_{d_2}-\mathbb{E}[x_{d_2}])], \\
\textrm{cum}(x_{d_1},x_{d_2},x_{d_3}) & = \mathbb{E}[(x_{d_1}-\mathbb{E}[x_{d_1}]) \notag\\
& (x_{d_2}-\mathbb{E}[x_{d_2}])(x_{d_3}-\mathbb{E}[x_{d_3}])].
\end{align*}

These empirical tensors $\hat{\bf{S}}$ and $\hat{\bf{T}}$ will
converge to
\begin{align}
\textbf{S} & = \sum_k s_k \textbf{a}_k \textbf{a}_k^T, \label{eq:S_decomposition_GP}\\
\textbf{T} & = \sum_k t_k \textbf{a}_k \otimes \textbf{a}_k \otimes \textbf{a}_k. \label{eq:T_decomposition_GP}
\end{align}
where $\textbf{a}_k$ is the $k^{th}$ column of $\textbf{A}$, $s_k = \textrm{var}(\alpha_k)$ and $t_k = \textrm{cum}(\alpha_k,\alpha_k,\alpha_k)$.

Note that we use the \emph{moments} for the Gaussian mixtures, while for the GP model we calculate the \emph{central moments}.

\section{Estimation of variance for Gaussian mixtures}
\label{appendix:gm_variance}
When applying the WTPM to Gaussian mixtures, some attention must be paid to correctly estimating the variance of the mixtures. If the estimate for $cov(\textbf{x},\textbf{x})$ is exact, its $D-K$ smallest eigenvalues are equal to $\sigma^2$, while all other eigenvalues are strictly larger than $\sigma^2$.
Therefore \citet{hsu2013learning} suggest using the smallest eigenvalue of $cov(\textbf{x},\textbf{x})$ as an estimate for the variance of the Gaussians. In practice, we find that the mean of the $D-K$ smallest eigenvalues yields better estimate, and that the quality of inference is very sensitive to this estimate.

Furthermore, poorly estimated moments lead to a large estimation error for $\sigma^2$, and therefore to perform the WTPM on Gaussian mixtures, we first calculate the variance of the $D_c-K$ smallest eigenvalues of $cov(\textbf{x}_c,\textbf{x}_c)$, where $\textbf{x}_c$ is the data vector containing only the $D_c$ complete dimensions.

Adopting the interpretation of weighting the moments estimates as a rescaling of the dimensions, such rescaling would deform the spherical Gaussian mixtures into elliptical Gaussians. In appendix \ref{appendix:elliptical_gaussians} we show that a weighting of the form given in \ref{eq:weighting_as_rescaling} naturally leads to the correct form of the moments, and therefore once $\sigma^2$ is calculated, no further modification is needed to apply the WTPM to Gaussian mixtures.

\section{Moments for elliptical Gaussian mixtures}
\label{appendix:elliptical_gaussians}

In this appendix we discuss the modifications to the moments estimates for elliptical Gaussian mixtures. We adopt the interpretation introduced earlier of viewing the weighting of the moments according to \ref{eq:weighting_as_rescaling} as a rescaling of the data, $x^*_{dn}=x_{dn}w_d$. Following a similar derivation to that presented in \cite{hsu2013learning} for non-spherical Gaussian mixtures, the estimates for the moments of the rescaled data are
\begin{align*}
\hat{S}^*_{d_1d_2} &= \mathbb{E}[x^*_{d_1}x^*_{d_2}] - \sigma^{*2}_{d_1}I \\
\hat{T}^*_{d_1d_2d_3} &= \mathbb{E}[x^*_{d_1}x^*_{d_2}x^*_{d_3}] - \sigma^{*2}_{d_1}\mathbb{E}[x^*_{d_1}]\delta_{d_2}\delta_{d_3} \\
&- \sigma^{*2}_{d_2}\delta_{d_1}\mathbb{E}[x^*_{d_2}]\delta_{d_3} - \sigma^{*2}_{d_3}\delta_{d_1}\delta_{d_2}\mathbb{E}[x^*_{d_3}],
\end{align*}

where $\sigma^{*2}_{d}$ is the standard deviation of the rescaled dimension $d$. The difficulty in applying the tensor decomposition method to non-spherical Gaussian mixtures lies in the fact that we don't know of a straight forward way compute $\sigma^{*2}_{d}$, whereas in the case of spherical mixtures ($\sigma^2_{d} = \sigma^2$ for all $d$), the variance is simply the smallest eigenvalue of the matrix $ \mathbb{E}[\bf{x \otimes x}] - \mathbb{E}[\bf{x}] \otimes \mathbb{E}[\bf{x}]$. However, if we know that the original data is generated by a spherical Gaussian mixture with standard deviation $\sigma^2$, the standard deviation of every dimension after rescaling by $w_d$ is $\sigma^{*2}_{d} = w_d^2\sigma^2$.

This implies that the rescaled moments can be written as
\begin{align*}
\hat{S}^*_{d_1d_2} &= w_{d_1}w_{d_2}\hat{S}_{d_1d_2}\\
\hat{T}^*_{d_1d_2d_3} &= w_{d_1}w_{d_2}w_{d_3}\hat{T}_{d_1d_2d_3}.
\end{align*}

This result, similar to equations \ref{eq:weighted_S} and \ref{eq:weighted_T} implies that the same weighting scheme used in our algorithm can be applied to Gaussian mixtures as well.

\section{Insensitivity to structure of topics}
\label{appendix:insensitivity_to_topics_structure}
\begin{figure}[ht]
\includegraphics[width=\linewidth]{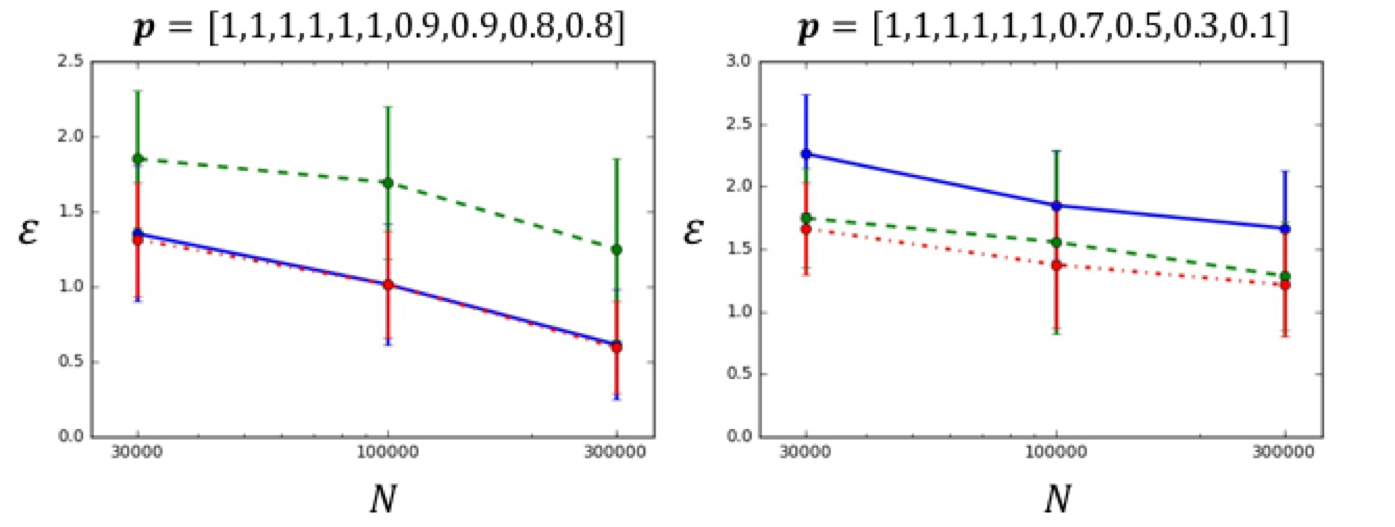}
\caption{Reconstruction error for the complete dimensions of synthetic data, $\varepsilon_c$, vs. $N$, where every incomplete dimension has a different probability, $p_d$, to be observed. A different topics matrix, $\bf{A}$, is sampled for each test. Each data point is an average over 25 runs.}
\label{fig:WTPM_demo_synthetic_random}
\end{figure}

In this appendix we demonstrate that our results are insensitive to the exact structure of the topics. In Figure~\ref{fig:WTPM_demo_synthetic_random} we show results similar to the results shown in Figure~\ref{fig:random_missing_data} for experiments performed with randomly generated topics. Each data point is an average over $25$ experiments, where for each experiment $\textbf{A}\in \mathbb{R}_+^{D\times K}$ was generated by sampling each of the $K=4$ columns in $\textbf{A}$ from $\text{Dir}(\textbf{1})$. The qualitative effect of a transition between the full dimensionality method to the partial dimensionality method being optimal is still observed, as well as the ability of the WTPM to perform at least as well as the better of the two methods for the entire range of parameter space studied.

\section{Optimal weights for minimization of inference error}
\label{appendix:optimal_weights}

\paragraph{Gamma-Poisson model}
We wish to find the weights $w_d$ which minimize the topics reconstruction error. Using results from \citet{anandkumar2012spectral}, \citet{podosinnikova2015rethinking} showed that the inference error is bounded by the sum of two contributions, one originating in the uncertainty in estimating $\hat{\textbf{S}}$, which scales as $\mathbb{E}\left[||\hat{\textbf{S}}-\textbf{S}||_F\right]/(\sigma_K(A)L)^2$, and one from $\hat{\textbf{T}}$, which scaling as $\mathbb{E}\left[||\hat{\textbf{T}}-\textbf{T}||_F\right]/(\sigma_K(A)L)^3$, where $\sigma_K(A)$ is the $K$-th largest singular value of $A$.

In the following we shall make the assumption that $D$ is significantly larger than $K$ and thus the singular values of the topics matrix will be well approximated by the dimensions with no missing values, that is, we may treat $\sigma_K(A)$ as constant with respect to our choice of weights.

We derive weights that minimize an upper bound of $\mathbb{E}\left[||\hat{\textbf{S}}-\textbf{S}||_F\right]/L^2$. The same set of weights $w_d\propto p_d$ approximately minimize $\mathbb{E}\left[||\hat{\textbf{T}}-\textbf{T}||_F\right]/L^3$, and the derivation is similar. Thus, we minimize an a quantity that is bounded away from the actual inference error by the sum of the Jensen gaps:
\begin{align*}
\mathbb{E}\left[||\hat{\textbf{S}}-\textbf{S}||_F\right] &< \sqrt{\mathbb{E}\left[||\hat{\textbf{S}}-\textbf{S}||_F^2\right] },\\
 \mathbb{E}\left[||\hat{\textbf{T}}-\textbf{T}||_F\right] &< \sqrt{\mathbb{E}\left[||\hat{\textbf{T}}-\textbf{T}||_F^2\right] }.
\end{align*}

We first observe that the weighting of $\hat{\textbf{S}}$ rescales the Frobenius error in the following way:
\small
\begin{align*}
\mathbb{E}\left[||\hat{\textbf{S}}^*-\textbf{S}^*||_F\right] & <\sqrt{\mathbb{E}\left[||\hat{\textbf{S}}^*-\textbf{S}^*||_F^2\right]} \\
& = \sqrt{\mathbb{E}\left[\sum_{d_1,d_2=1}^D(\hat{S}_{d_1d_2}^*-S_{d_1d_2}^*)^2\right]} \\
& = \sqrt{\mathbb{E}\left[\sum_{d_1,d_2=1}^D w_{d_1}^2 w_{d_2}^2 (\hat{S}_{d_1d_2}-S_{d_1d_2})^2\right]} \\
& = \sqrt{\sum_{d_1,d_2}^D w_{d_1}^2 w_{d_2}^2 \mathbb{E}\left[(\hat{S}_{d_1d_2}-S_{d_1d_2})^2\right]}
\end{align*}
\normalsize

We observe that the uncertainty in $\hat{\textbf{S}}$ scales as $\frac{1}{N_{d_1d_2}}$, where $N_{d_1d_2}$ is the number of times an estimate for $S_{d_1d_2}$ can be calculated from the data, i.e. the number of samples for which both $x_{d_1n}$ and $x_{d_2n}$ are observed - $N_{d_1d_2}=Np_{d_1}p_{d_2}$. Thus, we can choose $\gamma_{d_1, d_2}$ such that
\[
\left[\hat{\textbf{S}}-\textbf{S}\right]_{d_1, d_2} \leq \frac{\gamma_{d_1, d_2}}{Np_{d_1}p_{d_2}}.
\]
Note that $\gamma_{d_1, d_2}$ is independent of the weighting.
We can then write:
\begin{align*}
\mathbb{E}\left[||\hat{\textbf{S}}^*-\textbf{S}^*||_F\right] &\leq \sqrt{\sum_{d_1,d_2}^D \frac{\gamma_{d_1, d_2} w_{d_1}^2 w_{d_2}^2}{Np_{d_1}p_{d_2}}}\equiv E^*_S.
\end{align*}

The mean document length, $L = \mathbb{E}\left[\sum_{d_1}^D \mathbb{E}\left[X\right]_{d_1}\right]$, is also rescaled by the weights:
\begin{align*}
L^* & = \mathbb{E}\left[\sum_{d_1}^D \mathbb{E}\left[X\right]^*_{d_1}\right]\\
&= \mathbb{E}\left[\sum_{d_1}^D w_{d_1}\mathbb{E}\left[X\right]_{d_1}\right] \\
&= b\mathbf{w}^\top \mathbf{A}\mathbf{c}
\end{align*}

The goal is to minimize $E_S^*/(L^*)^2$, which is an upper bound for $\mathbb{E}\left[||\hat{\textbf{S}}^*-\textbf{S}^*||_F\right] / (L^*)^2$, over the closed unit hypercube in $\mathbb{R}^D$.

In general, we observe that $E_S^*/(L^*)^2$ (and respectively $E_T^*/(L^*)^3$) may not be convex in the choice of weights, thus an analytic derivation for the optima may not be feasible. However, under our existing assumptions, we see that $E_S^*/(L^*)^2$ (and respectively $E_T^*/(L^*)^3$) is coordinate-wise convex and hence we may seek local minima by computing the stationary points of $E_S^*/(L^*)^2$:

\begin{align*}
0 & = \frac{\partial}{\partial w_{d^*}}\left(\frac{E_S^*}{(L^*)^2}\right) \\
& =\frac{\partial}{\partial w_{d^*}}\left(\frac{\sqrt{\sum_{d_1,d_2}^D \frac{\gamma_{d_1, d_2} w_{d_1}^2 w_{d_2}^2}{Np_{d_1}p_{d_2}}}}{b\mathbf{w}^\top \mathbf{A}\mathbf{c}}\right).
\end{align*}
which simplifies to
\small
\begin{align*}
\frac{w_{d_*}}{p_{d_*}} = &\frac{[\mathbf{A}\mathbf{c}]_{d^*}}{b\mathbf{w}^\top \mathbf{A}\mathbf{c}} \left(\frac{ \displaystyle\sum_{d_1, d_2 \neq d^*} \frac{  \gamma_{d_1, d_2} w_{d_1}^2 w_{d_2}^2 }{Np_{d_1}p_{d_2}}} {\displaystyle\sum_{d_1 \neq d^*} \frac{  \gamma_{d_1, d_*} w^2_{d_1}}{Np_{d_1}}}\right) .
\end{align*}
\normalsize

For a high dimensional problem, choosing $D$ to be sufficiently large, we may consider the contributions from $w_{d_*}$ in RHS of the above negligible. In fact, for large $D$, the expression within the RHS brackets can be considered constant. Thus, we obtain the scaling $w_d^* \propto p_d^*$. 

Should we make an additional choice of a constant $\gamma$ such that
\[
\left[\hat{\textbf{S}}-\textbf{S}\right]_{d_1, d_2} \leq \frac{\gamma}{Np_{d_1}p_{d_2}},
\]
reflecting disregard for the structure in the data, we obtain the following expression for $E_S^*/(L^*)^2$
\[
E_S^*/(L^*)^2 = \frac{\sqrt{\frac{\gamma}{N}} \displaystyle \sum_{d} \frac{w^2_d}{\sqrt{p_d}}}{(b\mathbf{w}^\top \mathbf{A}\mathbf{c})^2}.
\]
The above formulation of $E_S^*/(L^*)^2$ is convex and yields a globally optimal weighting with the choice: $w_d\propto \sqrt{p_d}$. 

In experimentations we find that the two choices of weights, $w_d\propto \sqrt{p_d}$ and $w_d\propto p_d$ perform indistinguishably.

\paragraph{Generalization to other models}

In our approach, we expect the optimal weights to depend on the specific model used. Computing the optimal weights for different models requires complexity bounds results to determine how the inference error depends on the moments estimation error and any other parameters which might change with the weighting (an equivalent result to the $\varepsilon\sim\mathbb{E}[||\hat{\textbf{S}}-\textbf{S}||_F]/L^2$ scaling presented in the beginning of this appendix). Given these model dependent results, the optimal wrights for every moment can be easily computed by following the same straight-forward method used in this appendix, namely calculating the scaled inference errors and differentiating with respect to the weights.

\end{appendices}

\end{document}